\documentclass[letterpaper, 10 pt, conference]{ieeeconf}  

\IEEEoverridecommandlockouts                              

\usepackage[table]{xcolor}
\usepackage{svg}
\usepackage{graphicx} 
\usepackage{booktabs}
\usepackage{tabularx}
\usepackage{array}

\title{\LARGE \bf
UMR: Universal Manipulation Representation
}

\usepackage{multirow}   
\usepackage[font=footnotesize,labelfont=bf]{caption}
\usepackage{amsmath}
\usepackage{algorithm}
\usepackage{algorithmic}

\usepackage{cite}    
\makeatletter
\let\NAT@parse\undefined
\makeatother
\usepackage{hyperref}

\usepackage{subcaption}
\usepackage{dblfloatfix}
\usepackage{booktabs,makecell,graphicx,amssymb,pdflscape}
\definecolor{groupbg}{RGB}{231,222,241}\definecolor{highlightbg}{RGB}{235,235,235}
\newcommand{\cmark}{\ensuremath{\checkmark}}\newcommand{\xmark}{\ensuremath{\times}}

\author{
Song Liu$^{1,2,*}$, Linyi Li$^{1,2,*}$,
Yanshun Zhao$^{1}$, Rxuan Li$^{1}$, Xinrui Xu$^{1}$, Yi Ju$^{1}$,
Yahui Deng$^{1}$, Senge Zhang$^{1}$,\\ Guoyu Liu$^{1}$, Yixuan Li$^{1}$,Wuyang Zhang$^{1,2}$, Yao Li$^{1}$, Congcong Zhu$^{1,2,\dagger}$,
Jingrun Chen$^{1,\dagger}$
}

\begin{document}

\twocolumn[{
\renewcommand\twocolumn[1][]{#1}
\maketitle

\vspace{-2em}




\begin{center}
\refstepcounter{figure}\label{UMR_Overview}

\includegraphics[width=\textwidth]{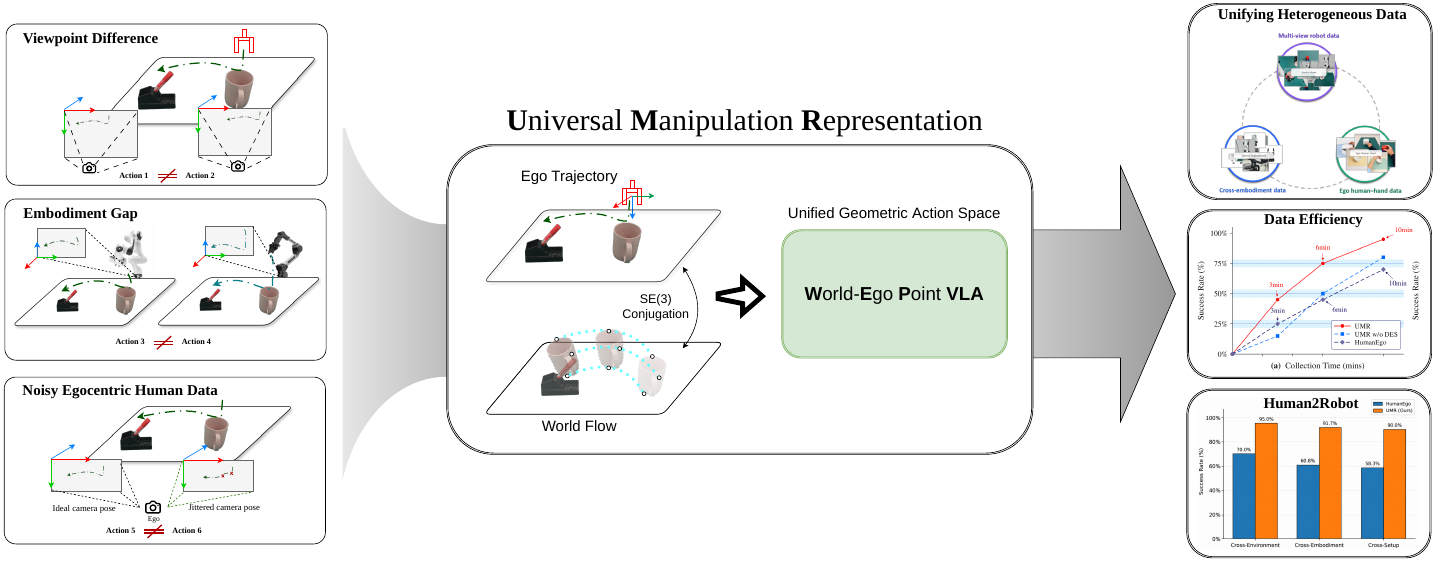}

\begin{flushleft}
\footnotesize
Fig.~\thefigure: UMR addresses the challenge of consistently representing and jointly learning from heterogeneous demonstrations through a unified World--Ego action representation, enabling WEPVLA to jointly learn in a unified action space, generalize across diverse conditions, and benefit from data scaling.
\end{flushleft}

\end{center}
}]

\begingroup
\renewcommand{\thefootnote}{}
\footnotetext{
\hspace{-1.5em}
$^{1}$University of Science and Technology of China, Hefei, China.
$^{2}$Suzhou Artificial Intelligence Laboratory, Suzhou, China.
$^{*}$Equal contribution.
$^{\dagger}$Corresponding author.
}
\endgroup


\begin{abstract}
General-purpose embodied manipulation hinges on a unified action representation that generalizes across embodiments and scales readily. 
Yet existing policies rely on embodiment-specific action spaces, making cross-embodiment demonstrations difficult to leverage at scale and limiting transfer to new embodiments and spatial variations.
To this end, we introduce Universal Manipulation Representation (UMR), a unified action representation that enables zero-shot skill transfer from human demonstrations to heterogeneous robots. UMR decomposes manipulation into two functionally distinct yet geometrically linked components: embodiment-agnostic World Flow, which describes task-relevant object motion in the world frame, and Ego Trajectory, which represents end-effector motion relative to the current pose.
We instantiate UMR as World--Ego Point VLA (WEPVLA), a compact 0.5B-parameter policy that learns in the unified geometric action space through a dual-stream Point Action Adapter and a unified Point Action Expert, with an $SE(3)$ conjugation coupling the two components.
To improve data efficiency, we
complement UMR with a Data-Efficient Strategy (DES) that diversifies object
configurations through stage-aware point-cloud editing while preserving
demonstrated contact geometry. 
In simulation, WEPVLA achieves average success rates of 97.5\% on LIBERO and 85.7\% on the 10-task RLBench benchmark. In real-world experiments, a single policy trained on human demonstrations augmented by DES transfers zero-shot to diverse deployment conditions. With about 10 minutes of collected human demonstrations per task and no robot demonstrations, it achieves 91.7\% average success across six evaluation settings, compared with 60.8\% for HumanEgo. Code and additional materials are available at
\color{red}\url{https://umr-wepvla.github.io/}.\end{abstract}

\begin{figure*}[t]
  \centering
  \scalebox{1}[1]{\includegraphics[width=1\textwidth]{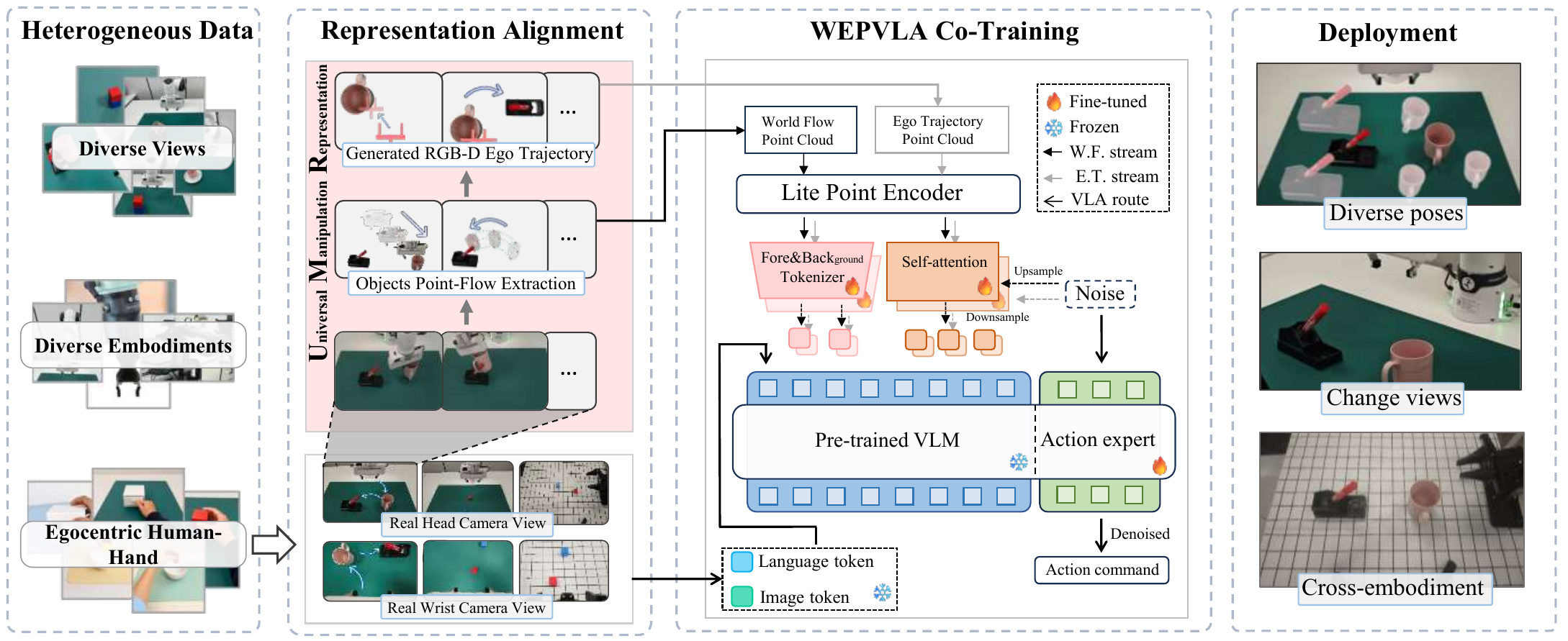}}
    \caption{\textbf{Overview of the UMR pipeline.} UMR reformulates the learning targets of heterogeneous manipulation data from different sources into a unified dual-stream representation consisting of World Flow and Ego Trajectory. Building on this representation, we develop WEPVLA with two dedicated branches to extract task-relevant and execution-relevant features, which are subsequently fused in a shared Action Expert for action generation. Experiments demonstrate strong generalization across robot embodiments, viewpoints, and spatial configurations.}
  
  \label{UMR_Pipeline}
\end{figure*}

\section{Introduction}

Scaling up data has become the prevailing route toward general-purpose embodied manipulation. Yet the effectiveness of data scaling rests on a more fundamental prerequisite: how to represent heterogeneous demonstrations. Existing policies operate in embodiment-specific joint or end-effector
spaces~\cite{OpenVLA,intelligence2025pi05visionlanguageactionmodelopenworld,smolvla,cen2025worldvla}, while available datasets encode actions under incompatible control semantics and coordinate frames~\cite{openx2023rtx,
wu2024robomind,khazatsky2024droid,liu2025fastumi100k,
liu2022hoi4d}. Consequently, demonstrations from humans and heterogeneous robots cannot be directly expressed in a shared action space and increasing data volume does not necessarily broaden the model's transferable capabilities. This calls for both a unified manipulation representation and an end-to-end policy that can jointly learn from heterogeneous demonstrations within this representation.

Recent 3D-aware policies use point clouds, depth, and poses to model spatially grounded manipulation~\cite{chen2026pointact,qu2025spatialvla, bridgevla,lin2026posevla}. Meanwhile, large-scale egocentric datasets establish natural human demonstrations as a scalable source of manipulation experience~\cite{grauman2021ego4d, grauman2023egoexo4d,hoque2025egodex,zheng2026egoscale, wu2026hug,li2026egowam}. However, transferring such demonstrations to robots still relies on specialized interfaces~\cite{chi2024umi, zhaxizhuoma2024fastumi,wei2026hifiumi}, motion retargeting~\cite{wu2026hug,zheng2026egoscale}, or embodiment-specific adaptation~\cite{li2026egowam}. These approaches therefore improve data acquisition without fully resolving the incompatibility between human and robot action spaces. 

Another line of work represents manipulation through object motion, 3D flow, or affordances, capturing physical effects that are less dependent on the acting embodiment~\cite{yuan2024generalflow, huang2026pointworld,tong2026pointaction,li2026egowam, han2026bridgeact}. Nevertheless, such world-level motion is not directly aligned with the local control interface of a robot and typically requires additional grasping, geometric fitting, action decoding, or control modules~\cite{yuan2024generalflow, huang2026pointworld,tong2026pointaction,han2026bridgeact}. While these designs enable effective transfer, their multi-stage pipelines often rely on task- or embodiment-specific interfaces between motion prediction and action generation. Together with inconsistent action representations, this separation hinders joint end-to-end learning from heterogeneous demonstrations and makes it harder to translate data scaling into generalization across tasks and embodiments.

To address these limitations, we introduce \textbf{Universal Manipulation Representation (UMR)}, a unified geometric action representation for end-to-end learning from heterogeneous demonstrations. UMR represents the same physical motion through two complementary geometric components. \textbf{World Flow} describes task-relevant motion as a structured $SE(3)$ transformation trajectory in the world frame. Under a fixed object--end-effector attachment, it removes the grasp transform and captures the same object motion across different demonstrators. \textbf{Ego Trajectory} expresses the corresponding end-effector motion relative to its current pose, removing dependence on the global coordinate frame while retaining a controller-aligned action representation. World Flow and Ego Trajectory are linked by $SE(3)$ conjugation and therefore constitute two coordinate views of the same underlying motion. This relation allows transferable task motion learned from one demonstrator to be expressed as an executable trajectory for another embodiment. Unlike unconstrained dense point flow, UMR represents motion through a compact and geometrically consistent $SE(3)$ trajectory. 

We instantiate UMR as \textbf{World--Ego Point VLA (WEPVLA)}, a point-cloud-based vision-language-action policy. WEPVLA employs a dual-stream \textbf{Point Action Adapter} to encode World Flow and Ego Trajectory under their respective geometric supervision, together with a shared \textbf{Point Action Expert} that jointly generates both action streams. During inference, the two streams are jointly denoised, while only the Ego Trajectory is executed by the robot. In this way, WEPVLA jointly learns transferable task-relevant object motion and executable actions in a fully end-to-end policy, without a separate task-specific stage for converting predicted world motion into robot actions. To make fuller use of the available demonstrations, we further introduce a \textbf{Data-Efficient Strategy (DES)} that diversifies relative object configurations while preserving demonstrated contact geometry. Experiments on LIBERO and RLBench demonstrate competitive manipulation performance, while real-world evaluations show zero-shot human-to-robot transfer across embodiments and scene configurations.

\section{Related Work}

\subsection{3D-Aware Vision-Language-Action Policies}

Pose-VLA separates universal 3D pose pretraining from robot-specific
action alignment~\cite{lin2026posevla}, while PointACT incorporates
hierarchical point-cloud features into action decoding through
multi-scale point--action interaction~\cite{chen2026pointact}.
Both enhance 3D reasoning but retain embodiment-specific action
representations. UMR instead represents a unified action space  through World Flow and Ego Trajectory.

\subsection{Egocentric and Robot-Free Human Demonstrations}

Robot-free human demonstrations provide scalable training data.
UMI~\cite{chi2024umi} and HiFi-UMI~\cite{wei2026hifiumi}
use handheld interfaces for data collection and robot deployment,
while HUG retargets human grasps to robot hands~\cite{wu2026hug}.
EgoWAM jointly learns world prediction and robot action from
egocentric data, using world prediction as auxiliary
supervision~\cite{li2026egowam}.
UMR instead unifies natural human and robot demonstrations in
a shared World--Ego action space, where object motion and
gripper execution are geometrically coupled.

\subsection{Motion Flow and Affordance Representations}

Flow- and affordance-based methods describe task effects
rather than embodiment-specific motion. BridgeACT predicts
\emph{where to grasp} and \emph{how to move}, converting these
predictions into robot actions through separate grasping
modules~\cite{han2026bridgeact}.

UMR couples embodiment-independent World Flow with executable
Ego Trajectory through an explicit $\mathrm{SE}(3)$ conjugation.
This formulation links object motion to executable action across varying scene configurations and grasp poses, enabling end-to-end cross-embodiment manipulation.

\section{Method} \label{sec:method} \subsection{Universal Manipulation Representation} 
UMR decomposes manipulation into two functionally distinct yet geometrically linked components: World Flow and Ego Trajectory, as shown in Fig.~\ref{UMR_Overview}.

\subsubsection{\textbf{World Flow}}
Let $E$, $O$, and $W$ denote the end-effector, manipulated object,
and fixed world frames, respectively. We use the notation
${}^{A}\mathbf{T}_{B}$ to denote the rigid transformation of frame
$B$ represented in frame $A$. Then the end-effector trajectory starting at $t$ with an action chunk of horizon $H$ can be defined as 
$\{{}^{W}\mathbf{T}_{E,t+k}\}_{k=0}^{H}$. Assume that the end-effector maintains rigid contact with the manipulated object within an action chunk, such that
\begin{equation}
    {}^{W}\mathbf{T}_{E,t}
    =
    {}^{W}\mathbf{T}_{O,t}\mathbf{G}, \quad{\mathbf{G}} = {}^{O}{\mathbf{T_E}},
\end{equation}
where $\mathbf{G}$ is a time-invariant transform within the action chunk. 
We further define the temporal relative transformation of frame $B$ from time
$t$ to $t+k$, represented in frame $A$, as
\begin{equation}
\label{ego}
\Delta^{A}\mathbf{T}_{B,t\rightarrow t+k}
=
{}^{A}\mathbf{T}_{B,t+k}
\left({}^{A}\mathbf{T}_{B,t}\right)^{-1},
\end{equation}
where $\Delta\mathbf{}^{W}\mathbf{T}_{E,t\rightarrow t+k}$ denotes the relative end-effector motion from time $t$ to time $t+k$ in frame $W$.
Therefore, the motion relative to the action chunk's initial time $t$  satisfies

\begin{equation}
\begin{aligned}
\Delta^{W}\mathbf{T}_{E,t\rightarrow t+k}
&=
{}^{W}\mathbf{T}_{E,t+k}
\left({}^{W}\mathbf{T}_{E,t}\right)^{-1}
\\&=
\left({}^{W}\mathbf{T}_{O,t+k}\mathbf{G}\right)
\left({}^{W}\mathbf{T}_{O,t}\mathbf{G}\right)^{-1}
\\&=
\Delta^{W}\mathbf{T}_{O,t\rightarrow t+k}
\end{aligned}.
\end{equation}
Since $\mathbf{G}$ is eliminated within the action chunk,
the end-effector trajectory implicitly encodes the object motion.
  Let i index the grasp configurations $\mathbf{G_i}$. World Flow can represent the same object motion across diverse grasp configurations.

\subsubsection{\textbf{Ego Trajectory}}
Although World Flow provides an embodiment-agnostic description of
task-relevant motion, it is represented in a fixed world frame and
cannot be directly executed across different embodiments. 
We therefore represent the same motion in the current end-effector frame as Ego Trajectory. Specifically, let $E_t$ denote the end-effector frame at time $t$, whose pose is ${}^{W}\mathbf{T}_{E,t}$. All future end-effector poses within the action chunk are aligned to $E_t$:
\begin{equation}\label{ego-trajecrory_define}
\left\{
{}^{E_t}\mathbf{T}_{E,t+k}
\right\}_{k=1}^{H}
=
\left\{
\left({}^{W}\mathbf{T}_{E,t}\right)^{-1}
{}^{W}\mathbf{T}_{E,t+k}
\right\}_{k=1}^{H}.
\end{equation}
Furthermore, Ego Trajectory and World Flow are linked by an $SE(3)$ conjugate transformation:
\begin{equation}
\begin{aligned}
\Delta^{E_t}\mathbf{T}_{E,t\rightarrow t+k}
&=
{}^{E_t}\mathbf{T}_{E,t+k}
\left({}^{E_t}\mathbf{T}_{E,t}\right)^{-1}
\\&=
\left({}^{W}\mathbf{T}_{E,t}\right)^{-1}
{}^{W}\mathbf{T}_{E,t+k}
\\&=
\left({}^{W}\mathbf{T}_{E,t}\right)^{-1}
\Delta^{W}\mathbf{T}_{E,t\rightarrow t+k}
\left({}^{W}\mathbf{T}_{E,t}\right)
\\&=
\left({}^{W}\mathbf{T}_{E,t}\right)^{-1}
\Delta^{W}\mathbf{T}_{O,t\rightarrow t+k}
\left({}^{W}\mathbf{T}_{E,t}\right)
\end{aligned}.
\end{equation}
Conversely, Ego Trajectory removes dependence on the global coordinate system and robot base configuration. The world-frame object motion can be converted into the egocentric executable representation.

Overall, Ego Trajectory and World Flow describe the same underlying
physical motion from different perspectives.
World Flow provides an embodiment-agnostic task-relevant motion prior,
while Ego Trajectory removes global coordinate dependency and provides
an executable representation for different embodiments.
Through grasp-dependent $SE(3)$ conjugation, Ego Trajectory preserves
the corresponding object motion and enables
cross-embodiment transfer.

\subsection{World--Ego Point VLA} 
We instantiate UMR as WEPVLA, a point-cloud VLA with a dual-stream Point Action Adapter and a unified Point Action Expert, as shown in Fig.~\ref{UMR_Pipeline}. Both streams are built on a shared Motion-Aware Segmentation module that
extracts interaction-relevant foreground.

\subsubsection{Motion-Aware Segmentation} Let $\mathbf{x}_{i,t}\in\mathcal{X}_t$ denote the $i$-th point of point cloud $\mathcal{X}_t$ at time $t$, which is expressed in the ${E_t}$ frame.  Specifically, let $\mathcal{K}_t\subseteq\{-T,\ldots,-1,1,\ldots,T\}$ denote a discrete set of temporal offsets sampled from the entire episode of length $T$.
We evaluate co-moving and world-static hypotheses for each temporal observation $k\in\mathcal{K}_t$: 
\begin{equation} \begin{aligned}
d_{i,t}^{\mathrm{mov}}&=\min_{\mathbf{y}\in\mathcal{X}_{t+k}} \|\mathbf{x}_{i,t}-\mathbf{y}\|_2
\\d_{i,t}^{\mathrm{sta}}&=\min_{\mathbf{y}\in\mathcal{X}_{t+k}} \|\left(\Delta^{E_t}\mathbf{T}_{E,t\rightarrow t+k}\right)^{-1}\mathbf{x}_{i,t}-\mathbf{y}\|_2
\end{aligned}. \end{equation}
The co-motion evidence is obtained from the normalized residual difference with motion-dependent weight $w_{t,k}$ via a robust temporal aggregation:
\begin{equation}
g_{i,t}=\sum_{k\in\mathcal{K}_t}\left[w_{t,k}\frac{d_{i,t}^{\mathrm{sta}}-d_{i,t}^{\mathrm{mov}}}{d_{i,t}^{\mathrm{sta}}+d_{i,t}^{\mathrm{mov}}+\epsilon}\right],
\end{equation}
where $\epsilon$ is a small constant introduced to avoid numerical
instability when the denominator approaches zero. The weight
$w_{t,k}$ accounts for the varying motion magnitude across temporal offsets $k$ at time $t$, and reflects the relative motion speed of the trajectory over the corresponding temporal interval.

We further fuse co-motion, approach, contact, and tool-sweep interaction cues into a soft foreground target:
\begin{equation}
q_{i,t}=1-\prod_{r\in\mathcal R}\left(1-e_{i,t}^{r}\right),\mathcal R=\{\mathrm{mov},\mathrm{app},\mathrm{con},\mathrm{sweep}\},
\end{equation}
where $e_{i,t}^r \in [0,1]$ denotes the foreground score of point $i$ under cue $r$ at time $t$, with higher scores indicating greater relevance to manipulation. Thus, the foreground includes both co-moving manipulated objects and static targets involved in tool interaction.

The resulting $q_{i,t}$ serves as offline supervision for the student PriorSeg network, implemented with a LitePT backbone~\cite{Yue_2026_CVPR} and a binary segmentation head. While the teacher exploits temporal point clouds and robot trajectories, the student predicts foreground from only the current XYZRGB point cloud, enabling single-frame inference.

\subsubsection{Point Modeling}
As shown in Fig.~\ref{UMR_Pipeline}, the segmented point cloud is processed by World--Ego streams before entering a shared Point Action Expert. The World stream represents observations and actions in the world frame $W$, while the Ego stream re-expresses them in the current end-effector frame $E_t$. Each stream first employs a Coarse-to-Fine Point Encoder to obtain fine-grained foreground tokens and coarse background tokens. 

The point-wise foreground features are then fused with flow-time and chunk-position-aware action tokens through bidirectional self-attention in the Point Action Adapter, producing point-wise action tokens. Finally, the dual-stream action and scene tokens are jointly processed with frozen VLM image and language tokens by the shared Point Action Expert for action prediction.

\subsubsection{Coupled Flow Matching}
Let $a^W,a^E\in\mathbb{R}^{H\times10}$ denote the World and Ego action chunks, where each action consists of a 9D pose (3D translation and 6D rotation~\cite{rot6d}) and a gripper state. Given $C_t={}^W\mathbf{T}_{E,t}\in{SE}(3)$, we sample an Ego prior $z^E$ and obtain its World counterpart by
\begin{equation}
z^W=\Phi_{C_t}(z^E),
\end{equation}
where $\Phi_{C_t}$ transforms the 9D pose by $C_t$ while preserving the gripper state, coupling the same physical prior across the two frames.
For $s\in\{E,W\}$ and flow time $\tau\in[0,1]$, we interpolate between the prior $z^s$ and target action $a^s$ as
\begin{equation}
x_\tau^s=(1-\tau)z^s+\tau a^s,\qquad u_\tau^s=a^s-z^s,
\end{equation}
where $x_\tau^s$ and $u_\tau^s$ denote the noisy action state and target velocity, respectively.
To bridge the two flows, let $B_\tau,W_\tau\in{SE}(3)$ be the pose components of $x_\tau^E$ and $x_\tau^W$. We align them into a common world-motion representation by
\begin{equation}
G_\tau^E=C_tB_\tau C_t^{-1},\qquad G_\tau^W=W_\tau C_t^{-1}.
\label{eq:conjugate_bridge}
\end{equation}
The resulting $G_\tau^E$ and $G_\tau^W$ condition the corresponding action tokens for joint denoising in the shared Point Action Expert.
The overall training objective is
\begin{equation}
\mathcal{L}
=\mathcal{L}_{E\text{-FM}}
+\lambda_W\mathcal{L}_{W\text{-FM}}
+\lambda_{\mathrm{seg}}\mathcal{L}_{\mathrm{PriorSeg}},
\end{equation}
where $\mathcal{L}_{E\text{-FM}}$ and $\mathcal{L}_{W\text{-FM}}$ supervise the Ego and World flow velocities, and $\mathcal{L}_{\mathrm{PriorSeg}}$ supervises manipulation-relevant foreground segmentation.
At inference, both flows are jointly denoised, while only the Ego action chunk is executed.

\subsection{Data-Efficient Strategy}
\label{DES}

To improve sample efficiency, we complement UMR with a \textbf{Data-Efficient Strategy
(DES)} based on stage-aware point-cloud editing. DES independently perturbs
task-relevant objects to diversify their relative spatial configurations, expanding the
coverage of object-object relations in the training data.

Given a demonstration $\tau$, rigid transformations are sampled for selected objects and applied to their point clouds. During interaction, the end-effector trajectory is transformed consistently with the manipulated object to preserve the demonstrated contact geometry:
\begin{equation}
{}^{W}\mathbf{T}_{E\prime,t}
=\mathbf{H}'_{\mathrm{obj}}\mathbf{H}_{\mathrm{obj}}^{-1}{}^{W}\mathbf{T}_{E,t},
\end{equation}
where $\mathbf{H}_{\mathrm{obj}}$ and $\mathbf{H}'_{\mathrm{obj}}$ denote the original and augmented object poses. Non-interaction trajectories are then regenerated to connect the edited interaction states, producing diverse and geometrically synthetic demonstrations.

\begin{figure}[htpb]
    \centering
    \begin{subfigure}[]{\columnwidth}
        \centering
        \includegraphics[width=\linewidth]{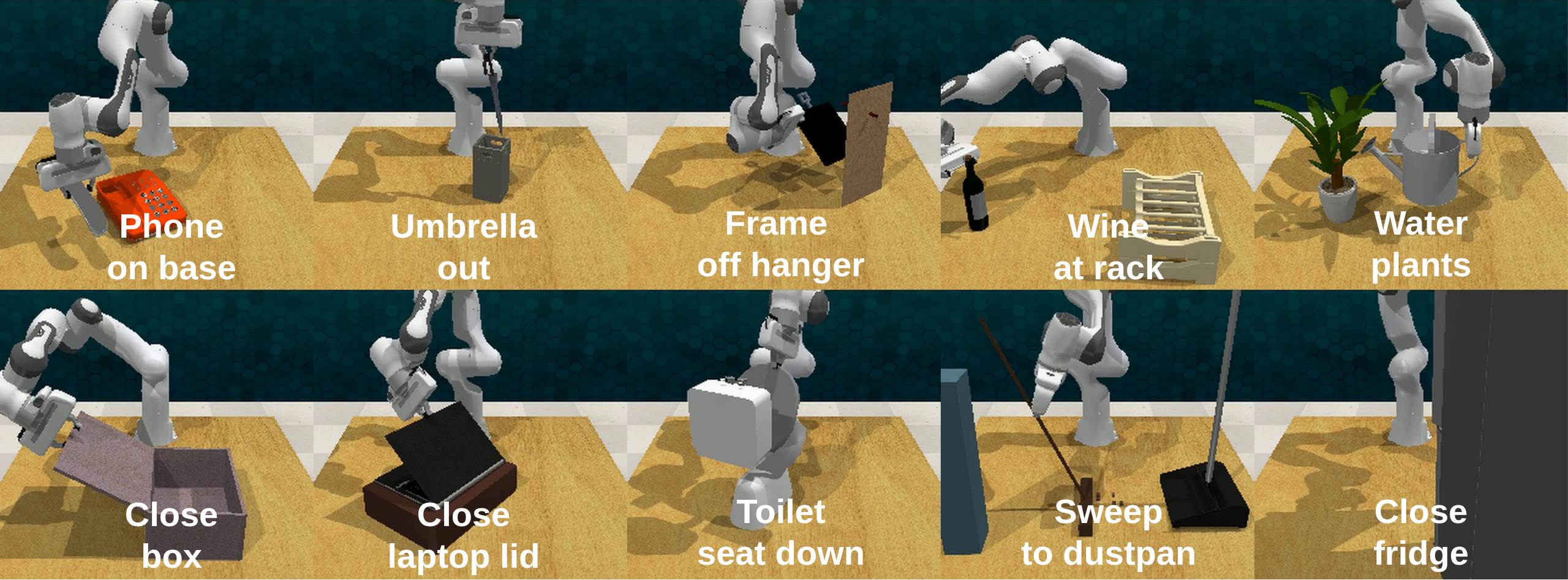}
        \caption{RLBench 10 tasks.}
        \label{rlbench_tasks}
    \end{subfigure}
    \vspace{2pt}
    \begin{subfigure}[]{\columnwidth}
        \centering
        \includegraphics[width=\linewidth]{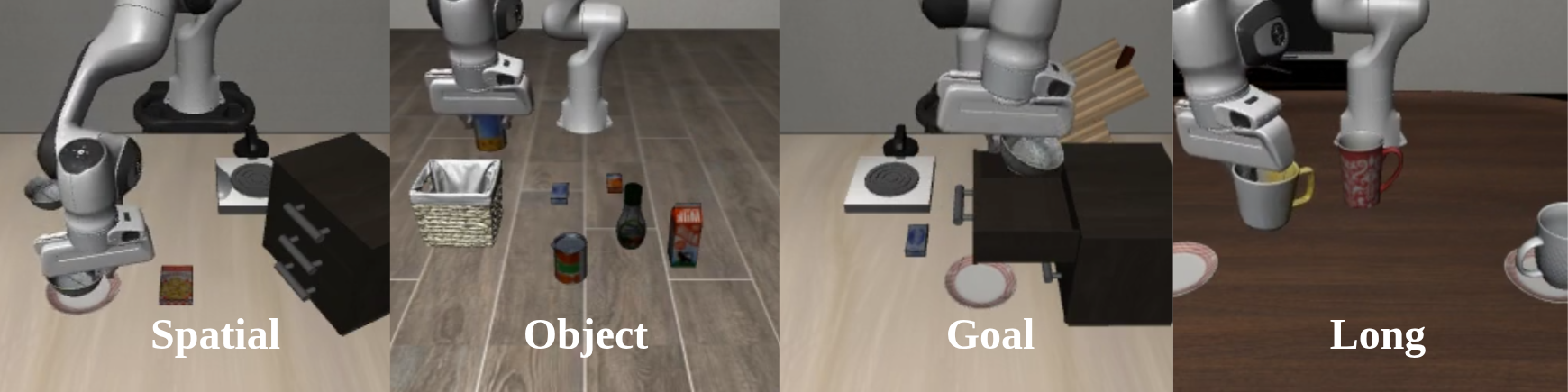}
        \caption{LIBERO 4 task suites.}
        \label{libero_tasks}
    \end{subfigure}
    \caption{\textbf{Overview of the simulated benchmarks.}
    (a) Ten RLBench tasks covering diverse manipulation scenarios that require object interaction and tool usage.
    (b) Four LIBERO suites, spanning tasks related to spatial relations, object interaction, goal specification, and long-horizon manipulation.}
    
    \label{simulation_benchmarks}
\end{figure}

\section{Experiments} \label{sec:experiments} We instantiate UMR as WEPVLA evaluated with respect to three questions: (1) whether UMR improves general manipulation performance in novel tasks, (2) whether human-only training transfers zero-shot across robot embodiments and scene configurations, and (3) whether the representation benefits from data scaling.

\subsection{Simulation Experiments}
We first evaluated WEPVLA on widely used benchmarks RLBench~\cite{james2020rlbench} and LIBERO~\cite{Libero} illustrated in Fig. \ref{simulation_benchmarks}.

\subsubsection{Experiment Setting}
\paragraph{RLBench}
Following PointACT~\cite{chen2026pointact}, we train a unified policy
on 10 tasks spanning articulated-object manipulation (close box,
fridge, and laptop lid; toilet seat down), grasp manipulation
(phone on base, stack wine, frame removal, and umbrella retrieval),
and tool use (water plants and sweep to dustpan).
We collect 100 expert demonstrations per task (1,000 total) using
the original language instructions. Front-view RGB images
($256\times256$) and point clouds are recorded at 20\,Hz, with
point clouds downsampled to 19,500 points. Following
HumanEgo~\cite{wang2026humanego}, we represent robot states as
virtual-gripper point clouds.

At inference, WEPVLA predicts a 32-step Ego Trajectory
(Eq.~\ref{ego-trajecrory_define}) and executes its first 28 actions.
The controller attempts Cartesian interpolation with inverse
kinematics, falling back to RRT-Connect~\cite{844730} with a
50\,ms sampling budget when infeasible.
We evaluate 100 episodes per task using the official success
criteria, with environment and policy-sampling seeds fixed to
100 and 20260801, respectively.

\paragraph{LIBERO}
We train a unified policy on all 40 tasks across LIBERO 4 suites, using 50 official demonstrations per task
(2,000 total). Replaying demonstrations from recorded simulator
states yields synchronized RGB-D observations and metric point clouds
at 20\,Hz, with controller-reconstructed targets as action labels.
We follow the RLBench training protocol with 10,000 input scene
points and adopt the evaluation configuration of OpenVLA~\cite{OpenVLA}.
Each task is evaluated on 50 initial states, with environment and
policy-noise seeds fixed to 7 and 0, respectively.

\begin{table}[htbp]
\centering
\caption{\textbf{Success rate (\%) on RLBench 10 tasks.}
We evaluate WEPVLA over 100 episodes per task.
Results for other methods are taken from \cite{chen2026pointact}.}
\label{RLBench_Eval}
\scriptsize
\setlength{\tabcolsep}{1.5pt}
\renewcommand{\arraystretch}{1.05}
\resizebox{\columnwidth}{!}{%
\begin{tabular}{lccccccccccc}
\toprule
&
\makecell{Close\\box} &
\makecell{Close\\laptop} &
\makecell{Toilet\\seat} &
\makecell{Sweep\\to dustpan} &
\makecell{Close\\fridge} &
\makecell{Phone\\base} &
\makecell{Umbrella\\out} &
\makecell{Frame\\off} &
\makecell{Wine\\rack} &
\makecell{Water\\plants} &
Mean \\
\midrule
ManipLLM (7B) ~\cite{Li_2024_CVPR}
& 50 & 80 & 40 & 20 & 80 & 35 & 10 & 25 & 15 & 20 & 38 \\
OpenVLA (7B) ~\cite{OpenVLA}
& 65 & 40 & 75 & 60 & 80 & 20 & 35 & 15 & 10 & 10 & 41 \\
$\pi_{0}$ (2.6B) ~\cite{BlackK-RSS-25}
& 90 & 60 & 100 & 30 & 90 & 25 & 35 & 75 & 5 & 45 & 55 \\
CogACT (7B) ~\cite{li2024cogact}
& 80 & 85 & 90 & 65 & 90 & 50 & 60 & 35 & 25 & 25 & 60 \\
HybridVLA (7B) ~\cite{liu2025hybridvla}
& 85 & 95 & 100 & 90 & 100 & 50 & 50 & 70 & 50 & 50 & 74 \\
EO1 (3B) ~\cite{qu2025eo}
& 97 & 99 & 100 & 95 & 83 & 46 & 76 & 40 & 61 & 35 & 73.2 \\
ACT3D ~\cite{pmlr-v229-gervet23a}
& 94 & 62 & 99 & 80 & 87 & 53 & 97 & 31 & 31 & 11 & 64.5 \\
PointACT (3B) ~\cite{chen2026pointact}
& 91 & 99 & 96 & 59 & 81 & 99 & 99 & 69 & 90 & 40 & 82.3 \\
\rowcolor{highlightbg}
WEPVLA (0.5B)
& 99 & 91 & 99 & 99 & 99 & 78 & 79 & 77 & 89 & 47 & \textbf{85.7} \\
\bottomrule
\end{tabular}%
}
\end{table}

\begin{table}[htbp]
\centering
\caption{Comparison of 2D VLAs and 3D-aware VLAs on LIBERO Bench.
“Pretrained” denotes whether the model is fully pretrained, and “train tasks”
denotes the number of LIBERO tasks the model is fine-tuned on.}
\label{LiberoBench_Eval}

\scriptsize
\setlength{\tabcolsep}{5pt}
\renewcommand{\arraystretch}{1}

\resizebox{0.49\textwidth}{!}{%
\begin{tabular}{lcccccccccc}
\toprule
&\makecell{Model\\size}
&\makecell{Wrist\\camera}
&\makecell{Pre-\\trained}
&\makecell{Frozen\\VLM}
&\makecell{Train\\tasks}
&Spatial&Object&Goal&Long&Avg.\\
\midrule

\rowcolor{groupbg}
\multicolumn{11}{c}{2D VLAs}\\
OpenVLA-OFT ~\cite{KimM1-RSS-25} &7B&\cmark&\cmark&\xmark&40&97.7&98.0&96.1&95.3&96.8\\
SmolVLA ~\cite{smolvla} &2B&\cmark&\cmark&\cmark&40&93&94&91&77&88.8\\
GR00T-N1.6 ~\cite{bjorck2025gr00t} &3B&\cmark&\cmark&\cmark&10&97.7&98.5&97.5&94.4&97.0\\
$\pi_{0.5}$ ~\cite{intelligence2025pi05visionlanguageactionmodelopenworld} &3B&\cmark&\cmark&\xmark&10&98.8&98.2&98.0&92.4&96.9\\
X-VLA ~\cite{zheng2026x} &0.9B&\cmark&\cmark&\xmark&40&98.2&98.6&97.8&\textbf{97.6}&98.1\\
MolmoAct2 ~\cite{fang2026molmoact2} &7B&\cmark&\cmark&\xmark&40&97.8&\textbf{100}&97.8&93.2&97.2\\
EO1 ~\cite{qu2025eo} &3B&\cmark&\cmark&\xmark&100&\textbf{99.7}&99.8&\textbf{99.2}&94.8&\textbf{98.4}\\

\midrule
\rowcolor{groupbg}
\multicolumn{11}{c}{3D-aware VLAs}\\
SpatialVLA ~\cite{qu2025spatialvla} &4B&\xmark&\cmark&\xmark&10&88.2&89.9&78.6&55.5&78.1\\
PointACT ~\cite{chen2026pointact} &3B&\xmark&\xmark&\cmark&10&97.4&99.6&96.2&90.6&96.0\\
\rowcolor{highlightbg}
WEPVLA (Ours)&0.5B&\xmark&\xmark&\cmark&40&\textbf{98.4}&\textbf{99.8}&\textbf{96.8}&\textbf{95.0}&\textbf{97.5}\\
\bottomrule
\end{tabular}%
}
\end{table}

\subsubsection{Comparison with State of the Art}

\paragraph{\textbf{State of the art on RLBench}}
As shown in Tab.~\ref{RLBench_Eval}, WEPVLA achieves the best average
success rate of 85.7\% on the 10-task RLBench benchmark, outperforming
the strongest baseline, PointACT, by 3.4 percentage points.
The improvement is particularly pronounced on \textit{Sweep to dustpan},
where WEPVLA increases the success rate from 59\% to 99\%.
This task requires coordinated tool use and precise reasoning over the
relative geometry between the tool, target object, and environment,
which is well aligned with UMR's explicit modeling of task-relevant object
motion through World Flow.
In contrast, WEPVLA performs worse than PointACT on
\textit{Phone on Base} and \textit{Umbrella Out} (78\% vs.\ 99\% and
79\% vs.\ 99\%, respectively).
We observe that these tasks frequently invoke configuration-space
planning near kinematic boundaries, producing trajectories with large
and redundant rotations, making Ego Trajectory
more difficult to learn.
\paragraph{\textbf{Smaller yet stronger on LIBERO}}
 WEPVLA achieves an average success rate of 97.5\% on LIBERO as shown in Tab.~\ref{LiberoBench_Eval}, while using
only 0.5B parameters and a single front-view observation.
Under the unified 40-task training setting, it outperforms SmolVLA
(88.8\%) and MolmoAct2 (97.2\%), despite being substantially smaller and
not requiring a wrist camera.
Among 3D-aware VLAs, WEPVLA achieves the best performance on all four
suites, improving over PointACT from 96.0\% to 97.5\% on average, with
the largest gain on the Long-Horizon suite (95.0\% vs.\ 90.6\%).
These results suggest that the gains of WEPVLA do not arise solely from
stronger 3D perception: by jointly modeling transferable World Flow and
executable Ego Trajectory in a shared policy, UMR provides a more
effective action representation for multi-task manipulation.
Overall, WEPVLA delivers competitive or state-of-the-art performance
with substantially fewer parameters, while maintaining a single policy
across all tasks.

\begin{figure}[htpb]
  \centering
\includegraphics[width=0.5\textwidth]{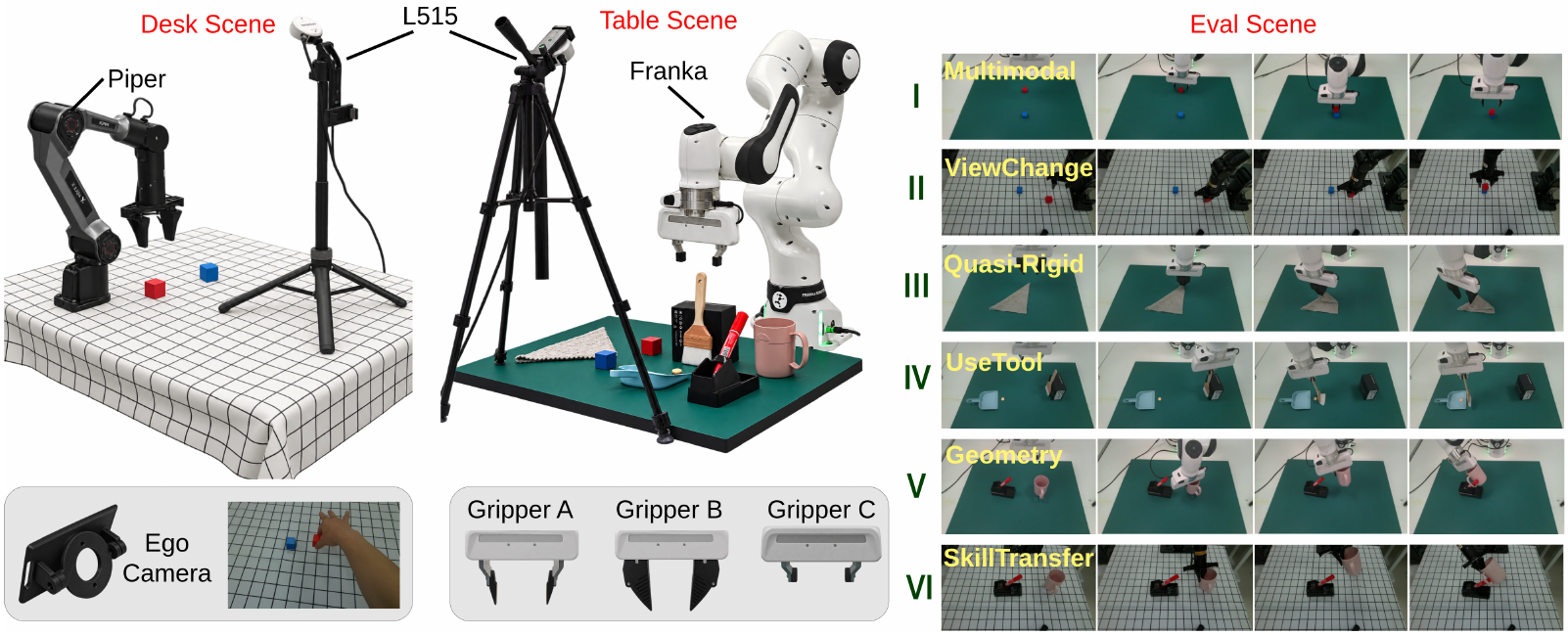}
    \caption{\textbf{Real-world setup and Eval Tasks.} 
    \textit{Left:} Experimental scene and hardware, manipulated objects, and training data source. 
    \textit{Right:} Examples of evaluation tasks under diverse deployment conditions.}
    \label{real_world_setting}
  \label{real_world_setting}
\end{figure}

\begin{table*}[t]
    \centering
    \begin{minipage}[t]{0.48\textwidth}
        \vspace{0pt}
        \centering
        \footnotesize
        \renewcommand{\arraystretch}{1}
        \setlength{\tabcolsep}{1.8pt}

        \captionof{table}{Real-world evaluation settings.}
        \label{tab_real_task_description}

        \begin{tabularx}{\linewidth}{
            >{\raggedright\arraybackslash}X       
            >{\centering\arraybackslash}m{0.4cm}        
            >{\centering\arraybackslash}X         
            >{\centering\arraybackslash}m{0.9cm}         
            >{\centering\arraybackslash}m{0.9cm}  
            >{\centering\arraybackslash}m{0.9cm}  
            >{\centering\arraybackslash}X         
        }
            \toprule
            Tasks & ID & Embodiments & Views & Scenes & Height
            & Disturbance \\

            \midrule
            \multirow{2}{*}{\textbf{CubeStacking}}
               & \uppercase\expandafter{\romannumeral1}  & Franka-A &\xmark & Table & \cmark & \xmark \\
               & \uppercase\expandafter{\romannumeral2}  & Piper    &  \cmark & Desk & \xmark & \xmark \\
            \textbf{TowelFolding}
               & \uppercase\expandafter{\romannumeral3}  & Franka-B & \xmark & Table & \xmark & \xmark \\
            \textbf{TrashSweep}
               & \uppercase\expandafter{\romannumeral4}    & Franka-C & \xmark & Table & \xmark & \cmark \\
            \multirow{2}{*}{\textbf{MugRack}}
               &  \uppercase\expandafter{\romannumeral5}   & Franka-C & \xmark & Table & \xmark & \xmark \\
               & \uppercase\expandafter{\romannumeral6}   & Piper    & \xmark & Desk & \cmark & \xmark \\
            \bottomrule

        \end{tabularx}
        
    \end{minipage}%
    \hfill
    \begin{minipage}[t]{0.48\textwidth}
        \vspace{0pt}
        \centering
        \footnotesize
        \renewcommand{\arraystretch}{1}
        \setlength{\tabcolsep}{2pt}

        \captionof{table}{Training data statistics.}
        \label{tab_training_datasets}

        \begin{tabularx}{\linewidth}{
            >{\raggedright\arraybackslash}X
            >{\centering\arraybackslash}X
            >{\centering\arraybackslash}X
            >{\centering\arraybackslash}X
            >{\centering\arraybackslash}X
            >{\centering\arraybackslash}X
        }
        \toprule
        \textbf{Source} & \textbf{Type} & \textbf{Tasks} &
        \textbf{Episodes} & \textbf{Views} & \textbf{Robots} \\
        \midrule
        
        \rowcolor{groupbg}
        \multicolumn{6}{c}{\small\textit{Public Datasets}} \\
        LIBERO~\cite{Libero}  & Sim  & 40  & 2000 & Exo       & 1 \\
        RLBench~\cite{james2020rlbench} & Sim  & 10  & 1000 & Exo       & 1 \\
        RH20T~\cite{10611615}   & Real & 146 & 8300 & Exo       & 4 \\
        
        \addlinespace[1pt]
        \rowcolor{groupbg}
        \multicolumn{6}{c}{\small\textit{Task Data}} \\
        Ours    & Real & 4   & 3000 & Exo / Ego & 0 \\
        
        \bottomrule
        \end{tabularx}

    \end{minipage}%
    
\end{table*}

\subsection{Real-World Experiments} 
We further evaluated the generalization of WEPVLA trained solely on human demonstrations in real-world environments under diverse cross-condition generalization.

\subsubsection{\textbf{Experiment setup}}
As shown in Fig.~\ref{real_world_setting}, we construct two real-world tabletop environments, a desk scene and a table scene, to evaluate WEPVLA under diverse deployment conditions. The setup includes two robot embodiments, Piper and Franka, each equipped with a parallel-jaw gripper, as well as a fixed Intel RealSense L515 RGB-D camera for scene observation. We further introduce three portable gripper configurations with different geometries to evaluate robustness to end-effector variations. Across the two scenes, seven everyday objects with diverse shapes, sizes, and physical properties are used for manipulation. Together, these configurations enable evaluation across changes in scene layout, viewpoint, robot embodiment, gripper geometry, and object properties.

\subsubsection{\textbf{Data Collection}}
We collect human demonstrations using a head-mounted Intel RealSense
L515 RGB-D camera, recording synchronized RGB-D frames, camera
intrinsics, timestamps, and hand motion. WiLoR~\cite{Potamias_2025_CVPR}
estimates 3D hand poses, from which the thumb and index finger are
mapped to a virtual parallel-jaw gripper to obtain robot-compatible
end-effector trajectories.

To compensate for head motion, we use the camera frame as the scene
reference for exocentric sequences. For egocentric sequences,
ORB-SLAM3~\cite{9440682} estimates camera motion, allowing reconstructed
point clouds and virtual-gripper trajectories to be transformed into
the first camera frame of each episode. The resulting metric-aligned
point clouds and trajectories serve as expert demonstrations.

\subsubsection{\textbf{Task Settings}}
We evaluate WEPVLA on four tasks as shown in Fig.~\ref{real_world_setting}:
\textit{CubeStacking} for grasping, relocation, and appearance reasoning;
\textit{MugRacking} for precise object--pose alignment;
\textit{TrashSweeping} for tool use; and \textit{TowelFolding} for deformable-object manipulation. Six evaluation settings test generalization across embodiments, viewpoints, scenes, heights, and disturbances as shown in Tab.~\ref{tab_real_task_description}.

For each task, we collect 100 human demonstrations in approximately
10 minutes: 50 nominal and 50 recovery demonstrations from
out-of-distribution initial states and recovery scenarios.
DES further generates 500 episodes per task, including full-stage
and stage-specific trajectories, yielding 600 training episodes
per task.

\begin{figure}[htpb]
    \centering
    \includegraphics[width=0.5\textwidth]{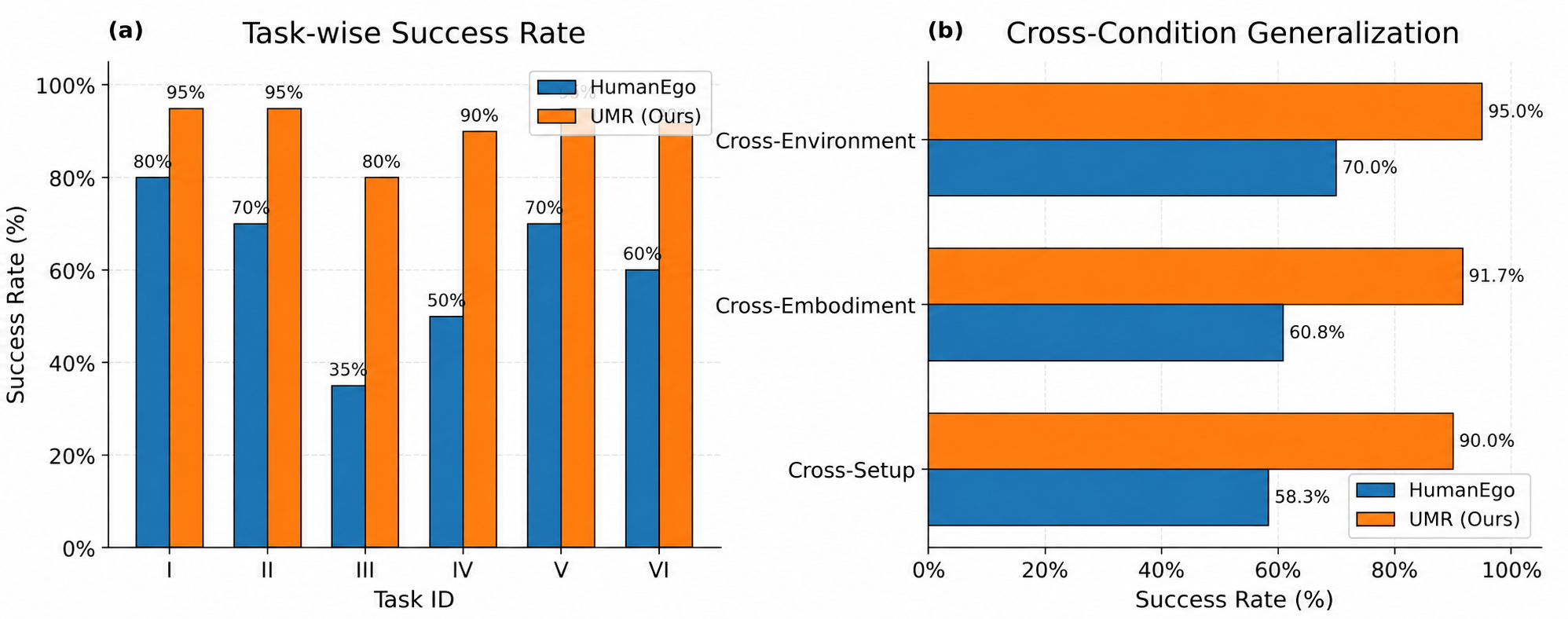}
    \caption{\textbf{Real-world comparison with HumanEgo.}
    Both methods are trained as unified policies on 10 minutes of human demonstrations per task.
    \textit{Left:} Success rates across six real-world evaluation (I--VI).
    \textit{Right:} Success rates aggregated by generalization condition.
    Cross-Embodiment includes Tasks I--VI, Cross-Environment includes
    Tasks I, II, V, and VI, and Cross-Setup includes Tasks I, III, and VI.
    UMR consistently outperforms HumanEgo across both individual tasks
    and aggregated generalization settings.}
    \label{UMR_vs_HumanEgo}
\end{figure}

\begin{figure*}[t]
    \centering
    \begin{subfigure}[t]{0.49\textwidth}
        \centering
        \scalebox{0.8}{\includegraphics[width=\linewidth]{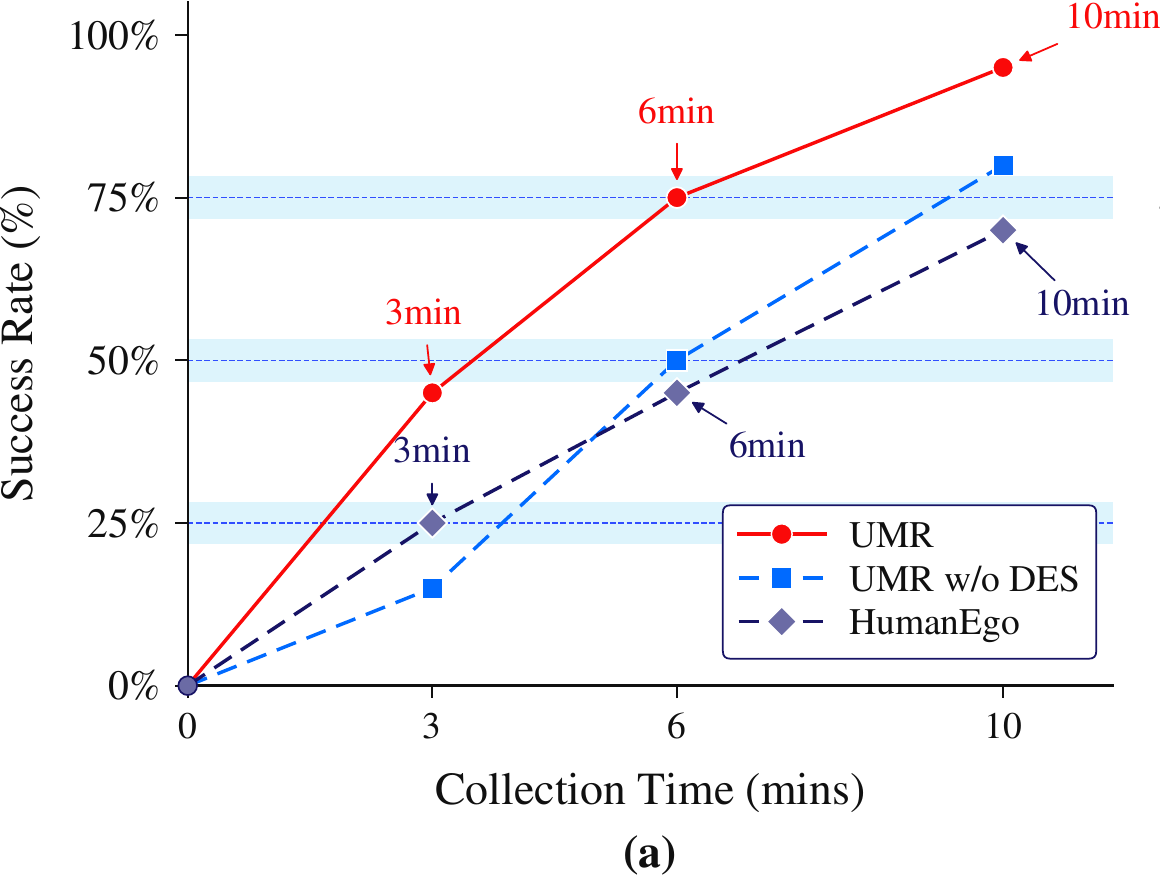}}
        \label{fig:umr-efficiency}
    \end{subfigure}
    \hfill
    \begin{subfigure}[t]{0.49\textwidth}
        \centering
        \scalebox{0.8}{\includegraphics[width=\linewidth]{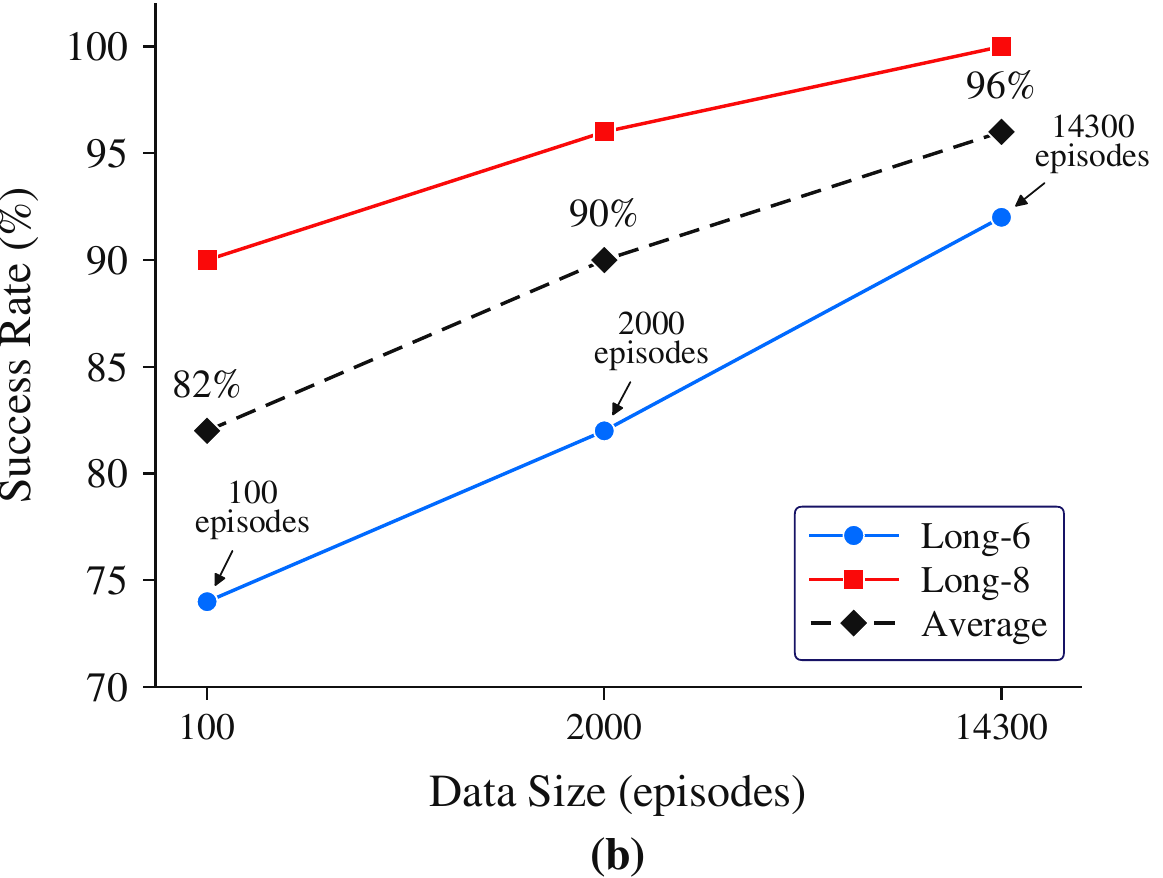}}
        \label{fig:umr-scaling}
    \end{subfigure}
    \caption{\textbf{Data efficiency and scaling of UMR.}
    \textbf{(a)} UMR achieves higher success rates with limited human
    demonstrations, while DES further improves data efficiency.
    \textbf{(b)} Scaling the training data improves performance on both
    long-horizon tasks, increasing average success from 82\% to 96\%.}
    \label{UMR_EfficiencyScaling_Resize}
\end{figure*}

\subsubsection{\textbf{One Policy, Zero-Shot across Tasks and Conditions}}

We evaluate whether a single policy trained solely on human demonstrations can generalize zero-shot across tasks, robot embodiments, and deployment conditions. This directly tests the core design of UMR: World Flow captures embodiment-agnostic task motion, while Ego Trajectory adapts it to the current end-effector for execution.

We compare WEPVLA with HumanEgo~\cite{wang2026humanego} under the same setting, where both methods are trained as unified multi-task policies using approximately 10 minutes of human demonstrations per task and no robot demonstrations. As shown in Fig.~\ref{UMR_vs_HumanEgo}, WEPVLA achieves an average success rate of \textbf{91.7\%}, substantially outperforming HumanEgo at \textbf{60.8\%} across Tasks I--VI. Aggregated over the cross-condition generalization, WEPVLA achieves success rates of \textbf{95.0\%}, \textbf{91.7\%}, and \textbf{90.0\%} under
cross-environment, cross-embodiment, and cross-setup generalization,
respectively, compared with \textbf{70.0\%}, \textbf{60.8\%}, and
\textbf{58.3\%} for HumanEgo.

These consistent gains across tasks and distribution shifts indicate that UMR captures transferable manipulation structure rather than condition-specific action mappings. Since this unified action representation is learned end to end within a single policy, it provides a natural basis for leveraging larger and more heterogeneous datasets; we therefore next investigate its scaling behavior with increasing data.

\subsection{Scaling Experiments}
We evaluate UMR from two complementary perspectives: \textit{data efficiency} and \textit{data scaling}. 

For data efficiency, we vary the human-demonstration budget from 3 to
10 minutes per task and compare UMR, UMR w/o DES, and HumanEgo under
the same evaluation protocol. DES augments limited demonstrations
through stage-aware point-cloud editing, perturbing task-relevant object
configurations while preserving the demonstrated contact geometry.
For data scaling, we evaluate Long-6 and Long-8, the two most
challenging long-horizon tasks for our model, to reduce ceiling effects
and better expose scaling behavior. We progressively expand the training
corpus from 100 task-specific LIBERO episodes (50 per task), to all
2,000 LIBERO episodes, and finally to 14,300 heterogeneous episodes
from the datasets summarized in Tab.~\ref{tab_training_datasets}, while keeping the model architecture and evaluation protocol fixed.

As shown in Fig.~\ref{UMR_EfficiencyScaling_Resize}(a), UMR achieves higher success rates with substantially less human demonstration data, while DES further improves performance in the low-data regime. Fig.~\ref{UMR_EfficiencyScaling_Resize}(b) further shows consistent gains as the training set grows: the average success rate on Long-6 and Long-8 increases from 82\% to 90\% and 96\% across the three data scales. Together, these results demonstrate that UMR is both data-efficient and able to benefit from data scaling.

\section{Ablation Studies}
\label{sec:ablation}

WEPVLA is built upon SmolVLA~\cite{smolvla}. To isolate the contributions of 3D perception, point--action modeling, and the World--Ego representation, we progressively augment the baseline with raw point clouds, motion-aware foreground segmentation, Point Modeling, Ego Trajectory, and World Flow, as summarized in Tab.~\ref{tab:libero-ablation}.

\subsection{From Raw 2D to 3D Perception}

Our evaluation of the official SmolVLA checkpoint yields an average
success rate of 81.65\%, compared with the reported 87.3\%.
Adding raw scene point clouds slightly reduces performance to 81.50\%,
indicating that unstructured 3D geometry alone is insufficient.
Motion-aware foreground segmentation raises success to 89.00\% by
focusing on interaction-relevant regions. Point Modeling further
improves it to 92.25\% by integrating point-wise geometry with
action tokens and multimodal features. These results highlight the
importance of both selecting relevant geometry and effectively
integrating it into action prediction.

\subsection{From 3D Perception to World--Ego Representation}

Introducing Ego Trajectory increases success to 96.40\% by expressing
actions relative to the current end-effector pose, reducing dependence
on global configurations and enabling continual re-anchoring during
execution. Adding World Flow further improves success to 97.50\%.
Its embodiment-agnostic task-motion supervision complements the
locally executable Ego Trajectory, helping the policy preserve
intended task effects across grasp poses and spatial configurations.

\par
\vspace{0.8\baselineskip}
\noindent
\begin{minipage}{0.48\textwidth}
    \centering
    \footnotesize
    \renewcommand{\arraystretch}{1}
    \setlength{\tabcolsep}{3.5pt}

    \captionof{table}{
        Success rates (\%) on LIBERO.
        Episodes denotes the total number of evaluation episodes for each method and the mean is calculated across all four suites. Policies are trained only in front view.
    }
    \label{tab:libero-ablation}

    \begin{tabularx}{\linewidth}{
        >{\raggedright\arraybackslash}p{2.4cm}
        *{6}{>{\centering\arraybackslash}c}
    }
        \toprule
        \textbf{Method}
        & Episodes
        & Spatial
        & Object
        & Goal
        & Long
        & \textbf{Mean} \\
        \midrule

        SmolVLA(Official)
        & 2000 & 82.00 & 92.00 & 85.20 & 67.40 & 81.65 \\

        + Point Cloud
        & 400 & 92.00 & 76.00 & 91.00 & 67.00 & 81.50 \\

        + Motion Seg.
        & 400 & 93.00 & 94.00 & 93.00 & 76.00 & 89.00 \\

        + Point Modeling
        & 400 & 97.00 & 97.00 & 92.00 & 83.00 & 92.25 \\

        + Ego Trajectory
        & 2000 & 97.60 & 99.10 & 94.70 & 94.00 & 96.40 \\

        + World Flow(\textbf{Ours})
        & 2000
        & \textbf{98.40}
        & \textbf{99.80}
        & \textbf{96.80}
        & \textbf{95.00}
        & \textbf{97.50} \\

        \bottomrule
    \end{tabularx}
\end{minipage}

\section{CONCLUSIONS}
In this work, we presented \textbf{UMR}, a unified action representation for learning manipulation from heterogeneous human and robot demonstrations. UMR couples embodiment-agnostic World Flow with locally executable Ego Trajectory through $\mathrm{SE}(3)$ conjugation, connecting transferable object motion with embodiment-conditioned execution. We instantiated UMR as \textbf{WEPVLA}, a compact 0.5B-parameter policy with a dual-stream Point Action Adapter and a shared Point Action Expert. The policy jointly predicts both components and executes only the Ego Trajectory. Motion-aware foreground segmentation focuses learning on interaction-relevant geometry, while DES enriches object configurations for data-efficient training. Simulation experiments show strong performance on RLBench and LIBERO, and real-world evaluations support zero-shot human-to-robot transfer across embodiments and scene configurations. Together, these results highlight the value of a shared geometric action space for transferable manipulation and provide a foundation for learning from heterogeneous demonstrations at scale.

\textbf{Limitations.}
The current implementation relies on precise RGB-D geometry and hand pose estimation, whose quality can deteriorate under severe occlusion or missing depth. Evaluation also focuses on tabletop manipulation with parallel-jaw grippers; broader validation with dexterous hands and more complex contact patterns remains necessary.

\textbf{Future Work.}
Future research will explore more robust 3D reconstruction from 2D images, larger-scale pretraining on diverse human and robot demonstrations, and extensions to dexterous and dual-arm manipulation, broadening the applicability of the unified representation.





\bibliographystyle{IEEEtran}
\bibliography{refs}

\end{document}